\pdfoutput=1
\documentclass[11pt]{article}

\usepackage[]{acl}

\usepackage{times}
\usepackage{latexsym}

\usepackage[T1]{fontenc}
\usepackage[utf8]{inputenc}

\usepackage{microtype}

\usepackage{graphicx}
\usepackage{multirow}
\usepackage{amsmath}
\usepackage{amsfonts}
\usepackage{booktabs}
\usepackage[english]{babel}
\usepackage{hyperref}
\addto\captionsenglish{%
}
\addto\extrasenglish{%
}

\title{Interpretability for Language Learners  \\ Using Example-Based Grammatical Error Correction}

\author{Masahiro Kaneko \quad Sho Takase \quad Ayana Niwa \quad Naoaki Okazaki\\
Tokyo Institute of Technology \\
\texttt{\{masahiro.kaneko, sho.takase, ayana.niwa\}@nlp.c.titech.ac.jp} \\ \texttt{okazaki@c.titech.ac.jp} 
}

\begin{document}
\maketitle
\begin{abstract}

Grammatical Error Correction (GEC) should focus not only on correction accuracy but also on the interpretability of the results for language learners.
However, existing neural-based GEC models mostly focus on improving accuracy, while their interpretability has not been explored.
Example-based methods are promising for improving interpretability, which use similar retrieved examples to generate corrections. 
Furthermore, examples are beneficial in language learning, helping learners to understand the basis for grammatically incorrect/correct texts and improve their confidence in writing.
Therefore, we hypothesized that incorporating an example-based method into GEC could improve interpretability and support language learners.
In this study, we introduce an Example-Based GEC (\textbf{EB-GEC}) that presents examples to language learners as a basis for correction result.
The examples consist of pairs of correct and incorrect sentences similar to a given input and its predicted correction.
Experiments demonstrate that the examples presented by EB-GEC help language learners decide whether to accept or refuse suggestions from the GEC output.
Furthermore, the experiments show that retrieved examples also improve the accuracy of corrections.

\end{abstract}

\section{Introduction}
\label{sec:intro}

Grammatical Error Correction (GEC) models, which generate grammatically correct texts from grammatically incorrect texts, are useful for language learners.
In GEC, various neural-based models have been proposed to improve the correction accuracy~\cite{yuan-briscoe-2016-grammatical,chollampatt-ng-2018-multilayer,junczys-dowmunt-etal-2018-approaching,zhao-etal-2019-improving,kaneko-etal-2020-encoder,omelianchuk-etal-2020-gector}.
However, the basis on which a neural GEC model makes corrections is generally uninterpretable to learners.
% In general, it is difficult for humans to interpret the basis used by neural networks for predictions~\cite{lei-etal-2016-rationalizing}.
Neural GEC models rarely address correction interpretability, leaving language learners with no explanation of the reason for a correction.

Interpretability plays a key role in educational scenarios~\cite{webb2020machine}.
In particular, presenting examples is shown to be effective in improving understanding.
Language learners acquire grammatical rules and vocabulary from examples~\cite{johns1994printout,mizumoto2015meta}.
Presenting examples of incorrect sentences together with correct ones improves the understanding of grammatical correctness as well as essay quality~\cite{arai-etal-2019-grammatical,arai2020example}.

\begin{figure}
\centering
\includegraphics[scale=0.26]{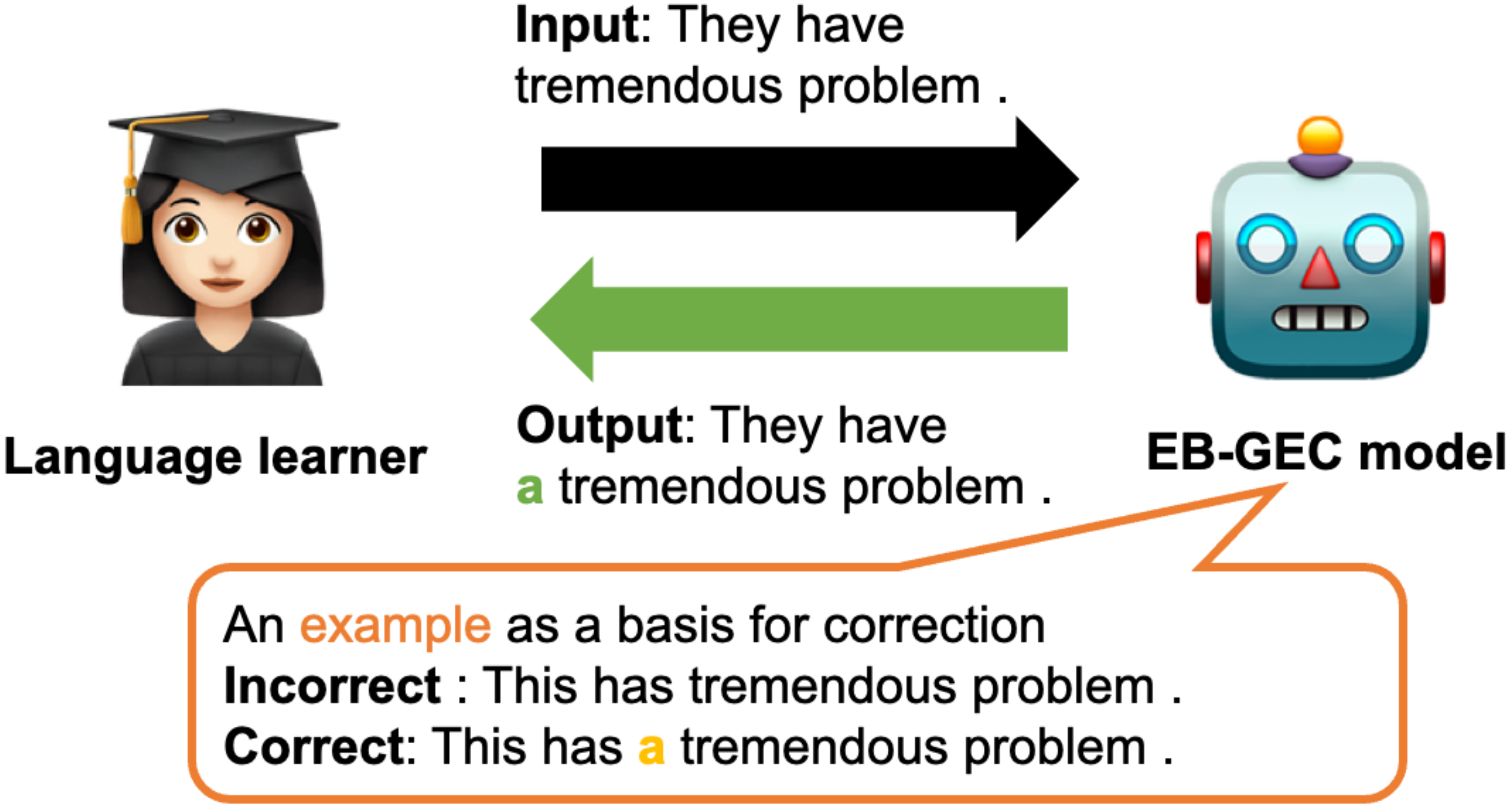}
\caption{EB-GEC presents not only a correction but also an example of why the GEC model suggested this correction.}
\label{fig:exgec}
\end{figure}

\begin{figure*}
\centering
\includegraphics[scale=0.54]{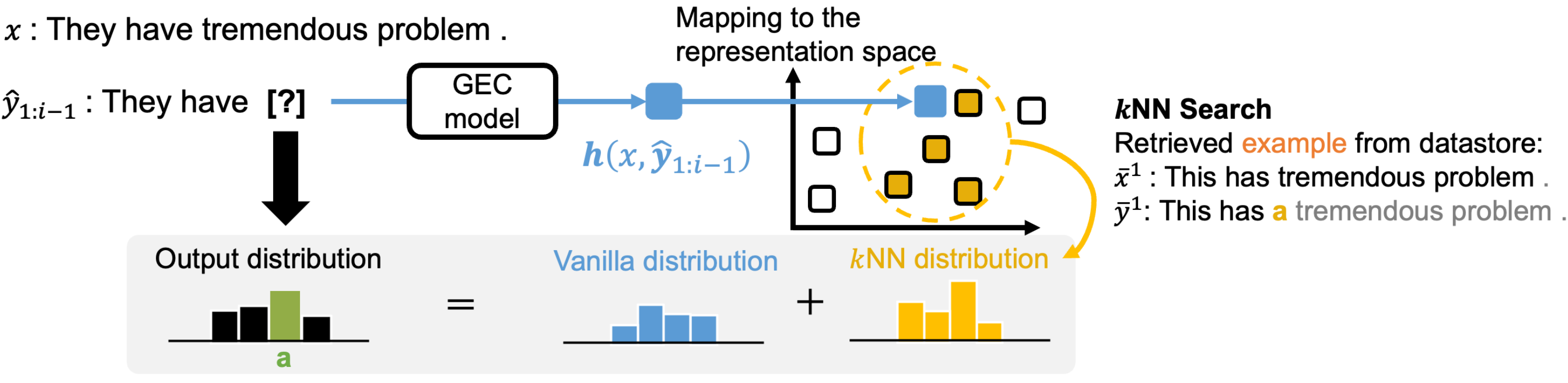}
\caption{An illustration of how EB-GEC chooses examples and predicts a correction. The model predicts a correction ``\textit{They have \underline{\hspace{0.3cm}/a} tremendous problem .}'' by using the example ``\textit{This has \underline{\hspace{0.3cm}/a} tremendous problem .}''
Hidden states of the decoder computed during the training phase are stored as keys, and tokens of the output sentences corresponding to the hidden states are stored as values. A hidden state of the decoder (blue box) at the time of inference is used as a query to search for $k$-neighbors (yellow box) of hidden states of the training data.
EB-GEC predicts a distribution of tokens for the correction from a combination of two distributions of tokens: a vanilla distribution computed by transforming the hidden state of the decoder; and a $k$NN distribution by the retrieved $k$-neighbors.
}
\label{fig:exgec_flow}
\end{figure*}

Recently, example-based methods have been applied to a wide range of natural language processing tasks to improve the interpretability of neural models, including machine translation~\cite{khandelwal2021nearest}, part-of-speech tagging~\cite{wiseman-stratos-2019-label}, and named entity recognition~\cite{ouchi-etal-2020-instance}.
These methods predict labels or tokens by considering the nearest neighbor examples retrieved by the representations of the model at the inference time.
\citet{khandelwal2021nearest} showed that in machine translation, examples close to a target sentence in the representation space of a decoder are useful for translating the source sentence.
Inspired by this, we hypothesized that examples corrected for similar reasons are distributed closely in the representation space.
Thus, we assume that neighbor examples can enhance the interpretability of the GEC model, allowing language learners to understand the reason for a correction and access its validity.

In this paper, we introduce an example-based GEC (\textbf{EB-GEC})\footnote{Our code is publicly available at \url{https://github.com/kanekomasahiro/eb-gec}} that corrects grammatical errors in an input text and provides examples for language learners explaining the reason for correction (\autoref{fig:exgec}).
%The \autoref{fig:exgec} shows the EB-GEC, where an example is a pair of incorrect and correct sentences such as ``\textit{This has \underline{\hspace{0.3cm}} /a tremendous problem .}'', and the example on which the correction is based is expected to be a similar example to the result of the correction.
%EB-GEC can use any method that would make the GEC model consider examples.
As shown in \autoref{fig:exgec_flow}, the core idea of EB-GEC is to unify the token prediction model for correction and the related example retrieval model from the supervision data into a single encoder-decoder model.
EB-GEC can present the reason for the correction, which we hope will help learners decide whether to accept or to refuse a given correction.

%Here, we note that EB-GEC does not assume a specific example-based model.
%We believe that presenting such examples related to the model input and output will support the learning process.

%The experimental results in \autoref{sec:result} show that EB-GEC can improve the interpretability while improving the accuracy by up to 2.3 points.
Experimental results show that EB-GEC predicts corrections more accurately than the vanilla GEC without examples on the three datasets and comparably on one dataset.
Experiments with human participants demonstrate that EB-GEC presents significantly more useful examples than the baseline methods of example retrieval~\cite{2008,yen-etal-2015-writeahead,arai2020example}.
%that the examples presented by EB-GEC provide explanations of correction results of the model and help users understand the correction results.
These results indicate that examples are useful not only to the GEC models but also to language learners.
This is the first study to demonstrate the benefits of examples themselves for real users, as existing studies~\cite{wiseman-stratos-2019-label,ouchi-etal-2020-instance,khandelwal2021nearest} only showed example utility for improving the task accuracy.
%We also found that the examples presented by EB-GEC often contain the same corrections and error types as the model output in \autoref{sec:cor_err_example}.
%On the other hand, the results of the analysis of the relationship between accuracy and interpretability in \autoref{sec:tradeoff} show that accuracy may deteriorate if GEC considers examples too much.

%However, existing GEC models focus more on accuracy and less on interpretability.
%Particularly, language learners may find it helpful to be presented the prediction basis of the model, thus improving the interpretability. 
%Learners then can decide whether to reflect corrections made by the model or not and may better understand the results of the corrections.
%Therefore, GEC models that satisfy not only high accuracy but also interpretability are needed.

%These neighbor examples may help the learner's acceptability and understanding of GEC outputs.

%In addition, we can design a GEC model so that it uses the retrieved examples in predicting a correction.
%This design may be more beneficial for language learners in terms of interpretability than retrieving examples found in the training data independently of the GEC model.
%Therefore, the presentation of examples related to the correction results of EB-GEC is considered to be helpful for language learning.

\section{EB-GEC}

%For the EB-GEC experiments, we used \textit{k}-Nearest-Neighbor Machine Translation~\cite[\textbf{\textit{k}NN-MT};][]{khandelwal2021nearest} as a GEC model.
%$k$NN-MT generates tokens by considering nearest neighbor examples using %representations from the GEC decoder at the time of inference.
%EB-GEC can use any method that would make the GEC model consider examples. Still, in this study, we use \textit{k}NN-MT due to its simplicity that does not require additional training for retrieving examples.

EB-GEC presents language learners with a correction and the related examples it used for generating the correction of the input sentence.
\textit{k}-Nearest-Neighbor Machine Translation~\cite[\textbf{\textit{k}NN-MT};][]{khandelwal2021nearest} was used as a base method to consider example in predicting corrections.
$k$NN-MT predicts tokens by considering the nearest neighbor examples based on representations from the decoder at the time of inference.
EB-GEC could use any method~\cite{gu2018search,zhang-etal-2018-guiding,NEURIPS2020_6b493230} to consider examples, but \textit{k}NN-MT was used in this study because it does not require additional training for example retrieval.

\autoref{fig:exgec_flow} shows how the EB-GEC retrieves examples using \textit{k}NN-MT.
EB-GEC performs inference using the softmax distribution of target tokens, referred to as vanilla distribution, hereafter, obtained from the encoder-decoder model and the distribution generated by the nearest neighbor examples. Nearest neighbor search is performed for a cache of examples indexed by the decoder hidden states on supervision data ($k$NN distribution).
EB-GEC can be adapted to any trained autoregressive encoder-decoder GEC model.
A detailed explanation of retrieving examples using  \textit{k}NN-MT is provided in Section \ref{sec:knn-mt}, and of presenting examples in Section \ref{sec:presnt_example}.

\subsection{Retrieving Examples Using \textit{k}NN-MT}
\label{sec:knn-mt}

Let $x=(x_1, ..., x_N)$ be an input sequence and $y=(y_1, ..., y_M)$ be an output sequence of the autoregressive encoder-decoder model.
Here, $N$ and $M$ are the lengths of the input and output sequences, respectively.

\paragraph{Vanilla Distribution.}
In a vanilla autoregressive encoder-decoder model, the distribution for $i$-th token $y_i$ of the output sequence is conditioned from the entire input sequence $x$ and previous output tokens $\hat{y}_{1:i-1}$, where $\hat{y}$ represents a sequence of generated tokens.
The probability distribution of the $i$-th token $p(y_{i}|x, \hat{y}_{1:i-1})$ is calculated by a linear translation to the decoder's hidden state $\boldsymbol{h}(x, \hat{y}_{1:i-1})$ followed by the softmax function.

\paragraph{Output Distribution.}
Let $p_{\rm EB}(y_{i}|x, \hat{y}_{1:i-1})$ denote the final probability distribution of tokens from EB-GEC.
We define $p_{\rm EB}(y_{i}|x, \hat{y}_{1:i-1})$ as a linear interpolation of the vanilla distribution $p(y_{i}|x, \hat{y}_{1:i-1})$ and $p_{\rm kNN}(y_{i}|x, \hat{y}_{1:i-1})$ (explained later), which is the distribution computed using the examples in the datastore,
\begin{align}
    \label{eq:final_dis}
    p_{\rm EB}(y_{i}|x, \hat{y}_{1:i-1}) =& \lambda p_{\rm kNN}(y_{i}|x, \hat{y}_{1:i-1}) \nonumber \\
    &+ (1 - \lambda) p(y_{i}|x, \hat{y}_{1:i-1}).
\end{align}
Here, $0 \leq \lambda \leq 1$ is an interpolation coefficient between the two distributions.
This interpolation also improves the output robustness when relevant examples are not found in the datastore.

\paragraph{Datastore.}
In the work of \citet{khandelwal2021nearest}, the $i$-th hidden state $\boldsymbol{h}(x,y_{1:i-1})$ of the decoder in the trained model was stored as a key, and the corresponding next token $y_i$ was stored as a value.
%$\tilde{\boldsymbol{h}}$ maps the input to the hidden state of the decoder, then the datastore
In order to present examples of incorrect/correct sentences, we stored a tuple of the token $y_i$, the incorrect input sentence $x$, and the correct output sentence $y$ as a value of the datastore.
Thus, we built key-value pairs ($\mathcal{K}$, $\mathcal{V}$) from all decoder timesteps for the entire training data $(\mathcal{X}, \mathcal{Y})$,
\begin{align}
    \label{eq:datastore}
    (\mathcal{K}, \mathcal{V}) = \{&(\boldsymbol{h}(x,y_{1:i-1}), (y_{i}, x, y)) \mid \nonumber \\
    & \forall y_i \in y, (x, y) \in (\mathcal{X}, \mathcal{Y})\} .
    %(\mathcal{K}, \mathcal{V}) = \{(\tilde{\boldsymbol{h}}_i, \tilde{y}_{i})\}^I_{i=1}
\end{align}
%where $\mathcal{H}$ is the set of all hidden states of decoder in the training data, and $\mathcal{Y}$ is the set of all target sentences for the training data.

\paragraph{\textit{k}NN Distribution.}
During inference, given a source $x$ as input, the model uses the $i$-th hidden state $\boldsymbol{h}(x,y_{1:i-1})$ of the decoder as the query to search for $k$-nearest neighbors,
\begin{align}
\mathcal{N} = \{(\boldsymbol{u}^{(j)}, (v^{(j)}, x^{(j)}, y^{(j)})) \in (\mathcal{K}, \mathcal{V})\}_{j=1}^{k} ,
\end{align}
where $\boldsymbol{u}^{(j)}$ ($j = 1, \dots, k$) are the $k$-nearest neighbors of the query $\boldsymbol{h}(x,y_{1:i-1})$ measured by squared $L^2$ distance.
The tuple $(v^{(j)}, x^{(j)}, y^{(j)})$ is the value associated with the key $\boldsymbol{u}^{(j)}$ in the datastore $(\mathcal{K}, \mathcal{V})$.
Then, the \textit{k}NN-MT aggregates the retrieved tokens to form a probability distribution $p_{\rm kNN}(y_{i}|x, \hat{y}_{1:i-1})$ with a softmax with temperature $T$ to the negative $L^2$ distances\footnote{In Equation \ref{eq:knn_distribution}, we do not use the input and output sentences in the value, and thus represent them as $\_$.},
\begin{align}
    \label{eq:knn_distribution}
    & p_{\rm kNN}(y_{i}|x, \hat{y}_{1:i-1}) \propto \nonumber \\
    & \sum_{(\boldsymbol{u}, (v, \_, \_)) \in \mathcal{N}} \mathbb{I}_{v = y_i} \exp\left(\frac{-\|\boldsymbol{u} - \boldsymbol{h}(x,\hat{y}_{1:i-1})\|}{T}\right) .
\end{align}
%where $d$ is a function that computes the squared $L^2$ distance of two vectors.
%$\hat{\boldsymbol{h}}_i$ represents the hidden state of the decoder computed using the generated tokens $\hat{y}_{1:i-1}$.

\subsection{Presenting Examples}
\label{sec:presnt_example}

We used a pair of incorrect and correct sentences stored in the value retrieved for the predicted token $\hat{y}_i$ as an example from the correction.
\autoref{fig:exgec} depicts an example where the retrieved value consists of the predicted token $v^{(j)} = \mbox{``\underline{\textit{a}}''}$ and the incorrect/correct sentences $x^{(j)}, y^{(j)}$ corresponding to ``\textit{This has \underline{\hspace{0.3cm}/a} tremendous problem .}''.
In this study, we presented examples for each edited token in an output.
For example, when an input or output is ``\textit{They have \underline{\hspace{0.3cm}/a} tremendous problem .}'', we presented examples for the edit ``\textit{\underline{\hspace{0.3cm}/a}}''.
%Similarly, if $\hat{y}_i$ is a token from the edit, then $x$ and $y$ of $\mathcal{N}_{\rm EB} = \{(\boldsymbol{q}^{(j)}, (v^{(j)}, x, y)) | j = 1, ..., k\}$ retrieved from $(\mathcal{K}_{\rm EB}, \mathcal{V}_{\rm EB})$ are used as an example.
To extract edit operations from an input/output pair, we aligned the tokens in input and output sentences by using the Gestalt pattern matching~\cite{ratcliff1988pattern}.

There are several ways to decide which examples should be presented to a language learner.
For instance, we could use all the examples in $k$-nearest neighbors $\mathcal{N}$ and possibly filter them with a threshold based on $L^2$ distance.
In this paper, we present an example incorrect/correct sentence pair that is the nearest to the query in $\mathcal{N}$, which is the most confident example estimated by the model.

\section{Experiments}

This section investigates the effectiveness of the examples via manual evaluation and accuracy on the GEC benchmark to show that the EB-GEC does, in fact, improve the interpretability without sacrificing accuracy.
We first describe the experimental setup and then report the results of the experiments.

\subsection{Datasets and Evaluation Metrics}
\label{sec:data}

We used the official datasets of BEA-2019 Shared Task~\cite{bryant-etal-2019-bea}, W\&I-train~\cite{Granger-1998,yannakoudakis-etal-2018}, NUCLE~\cite{dahlmeier-etal-2013-building}, FCE-train~\cite{yannakoudakis-etal-2011-new} and Lang-8~\cite{mizumoto-etal-2011-mining} as training data and W\&I-dev as development data.
We followed \citet{chollampatt-ng-2018-multilayer} to exclude sentence pairs in which the source and target sentences are identical from the training data.
The final number of sentence pairs in the training data was 0.6M.
We used this training data to create the EB-GEC datastore.
Note that the same amount of data is used by EB-GEC and the vanilla GEC model.

We used W\&I-test, CoNLL2014~\cite{ng-etal-2014-conll}, FCE-test, and JFLEG-test~\cite{napoles-sakaguchi-tetreault:2017:EACLshort} as test data.
To measure the accuracy of the GEC models, we used the evaluation metrics ERRANT~\cite{felice-etal-2016-automatic, bryant-etal-2017-automatic} for the W\&I-test and FCE-test, $\mathrm{M}^2$~\cite{dahlmeier-ng-2012-better} for CoNLL2014, and GLEU~\cite{napoles-EtAl:2015:ACL-IJCNLP} for the JFLEG-test.
$\mathrm{M}^2$ and ERRANT report F$_{0.5}$ values.

\subsection{Implementation Details of EB-GEC}

We used Transformer-big~\cite{NIPS2017_7181} as the GEC model.
Note that EB-GEC does not assume a specific autoregressive encoder-decoder model.
The beam search was performed with a beam width of 5.
We tokenized the data into subwords with a vocabulary size of 8,000 using BPE~\cite{sennrich-etal-2016-neural}.
%We tokenize the data into subwords with a vocabulary size of 8,000 using SentencePiece\footnote{\url{https://github.com/google/sentencepiece}}~\cite{kudo-2018-subword}.
%During training, we apply BPE-dropout~\cite{provilkov-etal-2020-bpe} to the source sentences with a dropout probability of 0.5.
The hyperparameters reported in \citet{NIPS2017_7181} were used, aside from the max epoch, which was set to 20.
In our experiments, we reported the average results of five GEC models trained using different random seeds.
We used four Tesla V100 GPUs for training.

We considered the \textit{k}NN and vanilla distributions equally, with $\lambda$ in Eq. \eqref{eq:final_dis} set to 0.5, to achieve both accuracy and interpretability.
Based on the development data results, the number of nearest neighbors $k$ was set to 16 and the softmax temperature $T$ to 1,000.
We used the final layer of the decoder feedforward network as the datastore key.
We used Faiss~\cite{8733051} with the same settings as \citet{khandelwal2021nearest} for fast nearest neighbor search in high-dimensional space.

\subsection{Human Evaluation Settings}

We assessed the interpretability by human evaluation based on~\citet{doshi2017towards}.
The human evaluation was performed to determine whether the examples improved user understanding and helped users to accept or refuse the GEC corrections.
To investigate the utility of the examples presented by EB-GEC, we examined the relative effectiveness of presenting examples in GEC as compared to providing none.
Moreover, we used two baseline methods for example selection, token-based retrieval and BERT-based retrieval.
Note that, unlike EB-GEC, token-based and BERT-based retrievals do not directly use the representations in the GEC model; in other words, these baselines perform the task of choosing examples independently of the GEC model.
In contrast, EB-GEC uses examples directly for generating an output.
EB-GEC was expected to provide examples more related to GEC input/output sentences than the baseline methods.

\paragraph{Token-based Retrieval.}
This baseline method retrieves examples from the training data where the corrections of the EB-GEC output match the corrections in the target sentence of the training data.
This is a similar method to the example search performed using surface matching~\cite{2008,yen-etal-2015-writeahead}.
If multiple sentences are found with matching tokens, an example is selected at random.
If the tokens do not match, this method cannot present any examples.

\paragraph{BERT-based Retrieval.}
This baseline method uses BERT\footnote{\url{https://huggingface.co/bert-base-cased}}~\cite{devlin-etal-2019-bert} to retrieve examples, considering the context of both the corrected sentence and example from the datastore.
This method corresponds to one based on context-aware example retrieval~\cite{arai2020example}.
In order to retrieve examples using BERT, we create a datastore,
\begin{align}
    \label{eq:bert_datastore}
    %& (\mathcal{K}_{\rm BERT}, \mathcal{V}_{\rm BERT}) \nonumber \\
    %& = \{(\tilde{\boldsymbol{b}}_i, (\tilde{y}_{i}, \tilde{x}, \tilde{y})) | (\boldsymbol{b}, x, y) \in (\mathcal{B}, \mathcal{X}, \mathcal{Y})\}
    (\mathcal{K}_{\rm BERT}, \mathcal{V}_{\rm BERT}) = \{&(\boldsymbol{e}(y_i), (y_{i}, x,y)) | \nonumber \\
    & \forall y_i \in y, (x,y) \in (\mathcal{X},\mathcal{Y})\} .
\end{align}
Here $\boldsymbol{e}(y_i)$ is the hidden state of the last layer of BERT for the token $y_i$ when the sentence $y$ is given without masking.
%and $\mathcal{B}$ is the set of all hidden states of BERT in the training data.
This method uses $\boldsymbol{e}(y_i)$ as a query for the model output sentence to then search the datastore for $k$ nearest neighbors.
%$\mathcal{N}_{\rm BERT} = \{(\boldsymbol{q}^{(j)}, (v^{(j)}, x, y)) | j = 1, ..., k\}$
\paragraph{}
%\vspace{0.2cm}

The input and output sentences of the GEC model and the examples from the baselines and EB-GEC were presented to the annotators with anonymized system names.
Annotators then decided whether the examples helped to interpret the GEC output or not, or whether they aided understanding of grammar and vocabulary.
The example sentence pair was labeled as 1 if it was ``useful for decision-making or understanding the correction'' and 0 otherwise.
We then computed scores for Token-based retrieval, BERT-based retrieval, and EB-GEC models by counting the number of sentences labeled with 1.
We confirm whether corrections with examples were more beneficial for learners than those without, and whether EB-GEC could present more valuable examples than those from the baselines.
Since it is not always the case that only corrected parts are helpful for learners~\cite{2008,yen-etal-2015-writeahead}, the uncorrected parts were also considered during annotation.
%\autoref{fig:he_example} shows an example of human evaluation.

% \begin{figure}
% \centering
% \includegraphics[scale=0.26]{figure/he_example.pdf}
% \caption{Human Evaluation of EB-GEC Example.}
% \label{fig:he_example}
% \end{figure}

We manually evaluated 990 examples provided by the three methods for 330 ungrammatical and grammatical sentence pairs randomly sampled from the W\&I-test, CoNLL2014, FCE-test, and JFLEG-test.
The human evaluation was performed by two annotators with CEFR\footnote{\url{https://www.cambridgeenglish.org/exams-and-tests/cefr}} proficiency level B and one annotator with level C\footnote{
They are not authors of this paper.
In this human evaluation, annotators with a middle and high proficiency level are selected in case annotators cannot understand errors/corrections and make a judgment whether the presented example is necessary or unnecessary.
Therefore, this study does not focus on whether annotators with lower proficiency levels find it helpful to see examples without explanation.}.
All three annotators evaluated different examples.

\subsection{Results}
\label{sec:result}

\begin{table}[t]
\centering
\small
\begin{tabular}{lccc}
\toprule
Method & Human evaluation score \\
\midrule
Token-based retrieval & 28.8 \\
BERT-based retrieval & 52.4 \\
EB-GEC & \textbf{68.8}$^{\dagger,\ddagger}$ \\
%Error type &  \\
\bottomrule
\end{tabular}
%\end{adjustbox}
\caption{Results of the human evaluation of the usefulness of Token-based retrieval, BERT-based retrieval and EB-GEC examples. Human evaluation score is the percentage of useful examples among those presented to the language learners. The $\dagger$ and $\ddagger$ indicate statistically significant differences of EB-GEC according to McNemar's test ($p < 0.05$) against Token-based retrieval and BERT-based retrieval, respectively.}
\label{tbl:human_eval}
\end{table}

\paragraph{Human Evaluation of Examples.}
\autoref{tbl:human_eval} shows the results of human evaluation of Token-based retrieval, BERT-based retrieval, and EB-GEC models.
The percentage of useful examples has increased significantly for EB-GEC compared to token-based and BERT-based retrieval baselines.
The percentage of useful examples from EB-GEC is greater than 50, which indicates that presenting examples is more useful than providing none.
This result is non-trivial because the percentage for token-based retrieval is only 28.8, which indicates that those presented examples were mostly useless.
Therefore, the examples for interpretability in EB-GEC support language learners' understanding and acceptance of the model output.
%As a result, it can be said that there are more useful examples for learners provided by EB-GEC than useless examples and can provide more useful examples compared to the baseline methods used in example retrieving systems.

\begin{table}[t]
\centering
\small
\begin{tabular}{lcccc}
\toprule
Method & W\&I & CoNLL2014 & FCE & JFLEG \\
\midrule
Vanilla GEC & 50.12 & 49.68 & 41.49 & \textbf{53.71} \\
EB-GEC & \textbf{52.45} & \textbf{50.51} & \textbf{43.00} & 53.46 \\
\bottomrule
\end{tabular}
%\end{adjustbox}
\caption{Accuracy of vanilla GEC model and EB-GEC model on W\&I, CoNLL2014, FCE and JFLEG test data.}
\label{tbl:gec}
\end{table}

\paragraph{GEC Accuracy.}
We examined the impact of using examples for the prediction of GEC accuracy.
\autoref{tbl:gec} shows the scores of the vanilla GEC and EB-GEC for the W\&I, CoNLL2014, FCE, and JFLEG test data.
The accuracy of EB-GEC is slightly lower for JFLEG but outperforms the vanilla GEC for W\&I, CoNLL2014, and FCE.
This indicates that the use of examples contributes to improving GEC model accuracy.

\begin{figure}[t]
\centering
\includegraphics[scale=0.4]{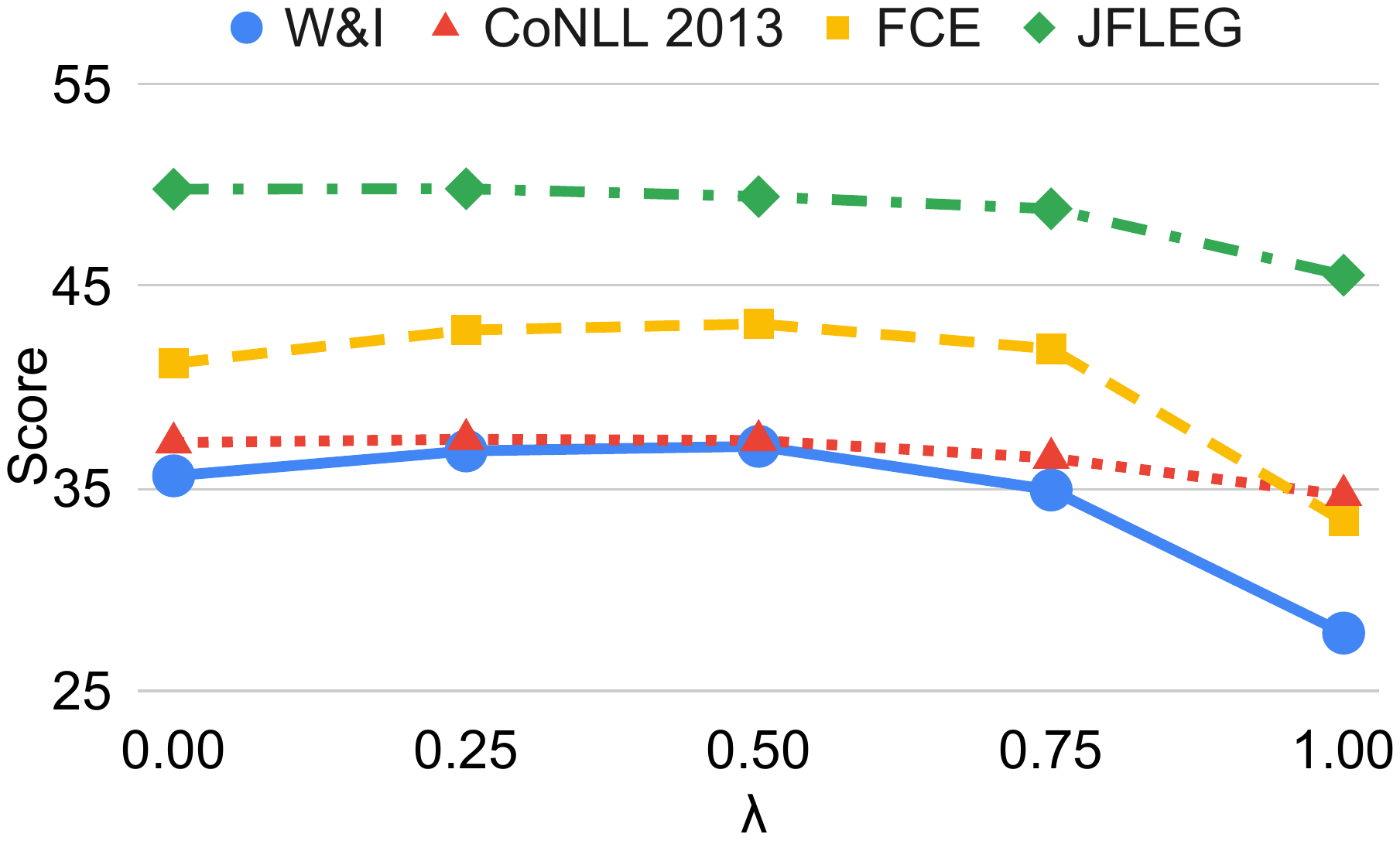}
\caption{Scores for each development data using different $\lambda$ values from 0 to 1 in increments of 0.25. The evaluation metrics for each data are the same as for the test data.}
\label{fig:lambda}
\end{figure}

\section{Analysis}

\subsection{Effect of $\lambda$}
\label{sec:tradeoff}

We analyzed the relationship between the interpolation coefficient $\lambda$ (in Equation \eqref{eq:final_dis}) and the GEC accuracy.
A smaller $\lambda$ value may reduce the interpretability as examples are not considered in prediction.
In contrast, a larger $\lambda$ value may reduce robustness, especially when relevant examples are not included in the datastore; the model must then generate corrections relying more on $k$NN examples, which may not be present in the datastore for some inputs.
% Therefore, we investigate the trade-off between accuracy and interpretability by adjusting the $\lambda$.

\autoref{fig:lambda} shows the accuracy of the GEC for each development data when the $\lambda$ is changed from 0 to 1 in increments of 0.25.
We found that when $\lambda$ was set to 1, the accuracy for all development datasets was lower than when $\lambda$ was set to 0.50 or less.
% This indicates that there is a trade-off between accuracy and interpretability in EB-GEC.
It is shown that the highest accuracy was obtained for $\lambda$ = 0.5, as this treats the vanilla output distribution and the output distribution equally.

\subsection{Matching Error Types of Model Outputs and Examples}
\label{sec:cor_err_example}

\begin{figure}[t]
\centering
\includegraphics[scale=0.48]{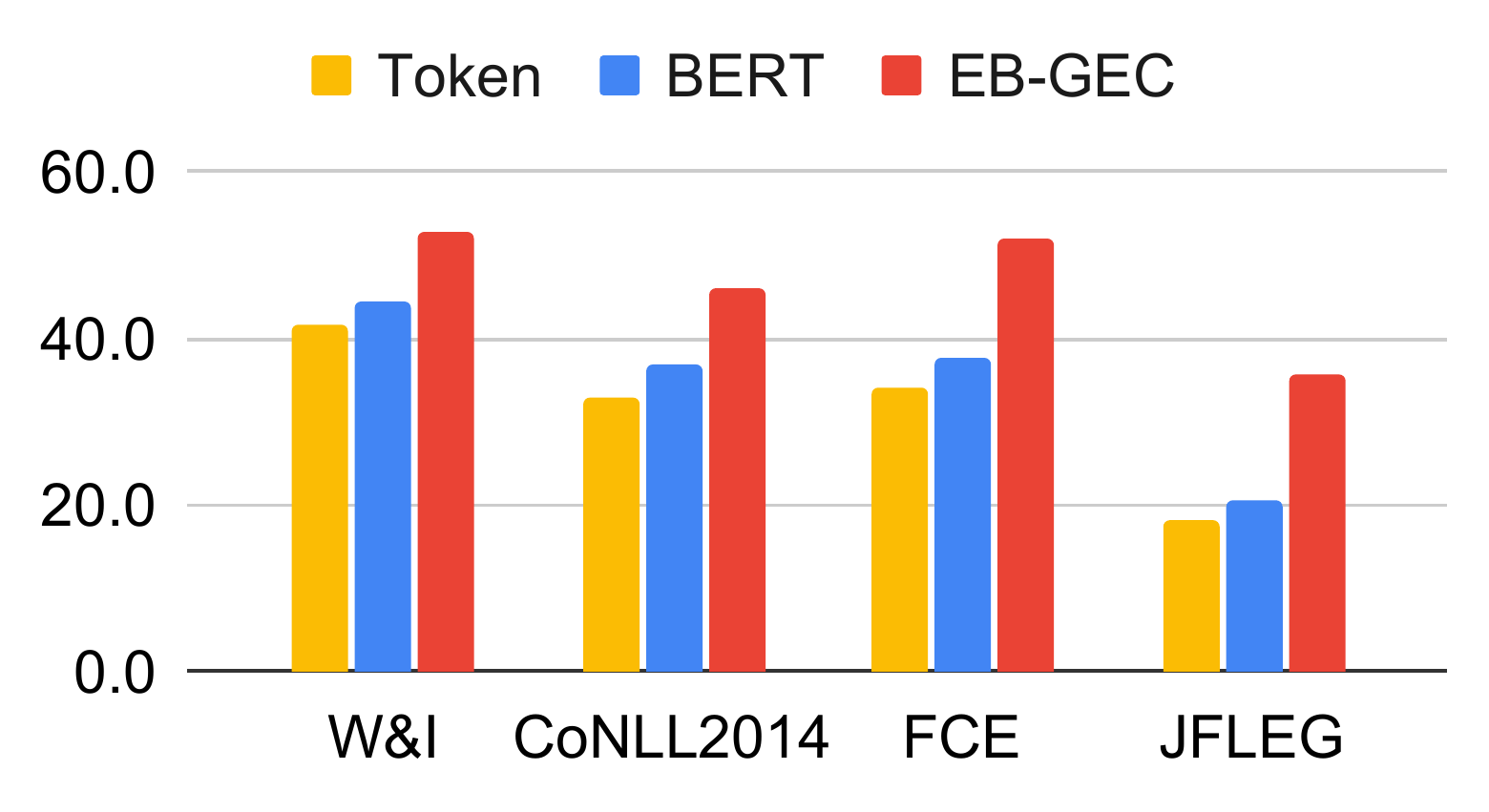}
\caption{Matching percentage of edits and error types in model outputs and examples.}
\label{fig:token_and_error_type}
\end{figure}

% To have more analysis of the examples, we perform an automatic evaluation of the examples on each test data.
In Section \ref{sec:intro}, we hypothesized that similar error-correcting examples are closely clustered in the representation space.
%Examples useful to language learners are often thought to contain error corrections similar to those in the GEC output.
Therefore, we investigated the agreement between the GEC output and the examples for edits and error types.
We extracted \textbf{edits} and their \textbf{error types}, which were automatically assigned by ERRANT~\cite{felice-etal-2016-automatic, bryant-etal-2017-automatic} for incorrect/correct sentence pairs.
For example, for a GEC input/output pair ``\textit{They have \underline{\hspace{0.3cm}/a} tremendous problem .}'', the example pair is ``\textit{This has \underline{\hspace{0.3cm}/a} tremendous problem .}'', its \textbf{edit} is ``\textit{\underline{\hspace{0.3cm}/a}}'' and the \textbf{error type} is the determiner error (\texttt{DET}). 
We calculated the matching percentage of the edits and error types for EB-GEC outputs and for the examples retrieved using EB-GEC to show their similarity.
In addition, we used token-based and BERT-based retrieval as comparison methods for obtaining examples relevant to EB-GEC outputs.

\autoref{fig:token_and_error_type} shows the matching percentage of edits and error types between the GEC outputs and the $k$-nearest neighbors examples.
First, we see that EB-GEC has the highest percentage for all test data.
This indicates that of the methods tested, EB-GEC retrieves the most relevant examples.
This trend is consistent with the human evaluation results.
Furthermore, we see that EB-GEC has a lower percentage on JFLEG compared to those on W\&I, CoNLL2014, and FCE.
This corroborates the results of \autoref{tbl:gec}, which suggests that the accuracy of GEC improved further when examples more relevant to the corrections could be retrieved.
%On the other hand, the percentage of matched cases was generally low compared to the percentage of cases that were useful in the manual evaluation shown in the figure.
%This may be because there are cases in which the correction pairs and error types do not match, such as when the correction does not reflect the correction due to a mismatch between the correction and the case, or when the correction pairs and error types do not match, such as when learning from a part that is not incorrect as in ordinary example searches.

\subsection{EB-GEC and Error Types}

\begin{table}[t]
\centering
\small
\begin{tabular}{lcccc}
\toprule
Error type & Freq. & Vanilla GEC & EB-GEC & Diff. \\
\midrule
\texttt{PREP} & 115K & 40.9 & 44.6 & 3.7 \\
\texttt{PUNCT} & 98K & 33.5 & 37.0 & 3.5 \\
\texttt{DET} & 171K & 46.6 & 49.8 & 3.2 \\
\midrule
\texttt{ADJ:FORM} & 2K & 54.5 & 38.4 & -16.08 \\
\texttt{ADJ} & 21K & 17.0 & 14.5 & -2.42 \\
\texttt{SPELL} & 72K & 68.6 & 66.8 & -1.87 \\
%Method & ADJ:FORM & NOUN:POSS & ADU & VERB & PREP & PART \\
%\midrule
%Original & \\
%EB-GEC & \\
\bottomrule
\end{tabular}
%\end{adjustbox}
\caption{The error types with the highest and the lowest EB-GEC accuracy compared to vanilla GEC on FCE-test based on Diff. column.
Freq. column is the frequency of the error type in the datastore.}
\label{tbl:error_type}
\end{table}

\begin{table*}[t]
\centering
\small
\begin{tabular}{lclc}
\toprule
& Error type & Error-correction pair & Label \\
\midrule
Input/Output & \texttt{PREP} & \textit{You will be able to buy them \textbf{\underline{in/at}} \underline{\hspace{0.3cm}/a} reasonable price .} & -  \\
\multirow{2}{*}{Token-based retrieval} & \multirow{2}{*}{\texttt{PREP}} & \textit{Naturally , it 's easier to get a job \underline{then/when} you \underline{were/are} good \textbf{\underline{in/at}}} & \multirow{2}{*}{0} \\
& & \textit{foreign \underline{languagers/languages} or computers .} & \\
BERT-based retrieval & \texttt{PREP} & \textit{I could purchase them \textbf{\underline{in/at}} reasonable \underline{price/prices}} . & 1 \\
EB-GEC & \texttt{PREP} & \textit{I could purchase them \textbf{\underline{in/at}} reasonable \underline{price/prices} .} & 1 \\
\midrule
Input/Output & \texttt{PUNCT} & \textit{\underline{for/For} example \textbf{\underline{\hspace{0.3cm}/,}} a reasercher that wants to be successfull must take risk .} & -  \\
Token-based retrieval & \texttt{PUNCT} & \textit{Today \textbf{\underline{\hspace{0.3cm}/,}} we \underline{first/\hspace{0.3cm}} met for \underline{\hspace{0.3cm}/the first time in} about four weeks .} & 0 \\
BERT-based retrieval & \texttt{PUNCT} & \textit{\underline{for/For} example \textbf{\underline{\hspace{0.3cm}/,}} a kid named Michael .} & 1 \\
EB-GEC & \texttt{PUNCT} & \textit{\underline{for/For} example \textbf{\underline{\hspace{0.3cm}/,}} a kid named Michael .} & 1 \\
\midrule
Input/Output & \texttt{DET} & \textit{Apart from that \underline{\hspace{0.3cm}/,} it takes \textbf{\underline{\hspace{0.3cm}/a}} long time to go somewhere .} & -  \\
Token-based retrieval & \texttt{DET} & \textit{If you have enough time , I recommend \textbf{\underline{\hspace{0.3cm}/a}} bus trip .} & 0 \\
BERT-based retrieval & \texttt{PREP} & \textit{However , it will take \textbf{\underline{for/\hspace{0.3cm}}} a long time to go abroad in my company .} & 0 \\
EB-GEC & \texttt{DET} & \textit{\underline{So/Because of that ,} it takes \textbf{\underline{\hspace{0.3cm}/a}} long time to write my \underline{journal/entries} .} & 1 \\
% \midrule
% Input/Output & \texttt{ADJ:FORM} & & -  \\
% Token-based retrieva &  \\
% BERT-based retrieval &  \\
% EB-GEC &  \\
% \midrule
% Input/Output & \texttt{ADJ} & The \underline{mainly/main} reasons are \textbf{\underline{such/\hspace{0.3cm}}} as \underline{\hspace{0.3cm}/the} \underline{following/follows} \underline{./:} & -  \\
% Token-based retrieva &  \\
% BERT-based retrieval & &   \\
% EB-GEC & & & 0 \\
% \midrule
% Input/Output & \texttt{SPELL} & & -  \\
% Token-based retrieva &  \\
% BERT-based retrieval &  \\
% EB-GEC &  \\
\bottomrule
\end{tabular}
%\end{adjustbox}
\caption{Examples retrieved by Token-based retrieval, BERT-based retrieval, and EB-GEC for input/output, and the  human evaluation labels.
\underline{Underlines} indicate error-correction pairs in the sentences.
\textbf{Bold} indicates the edit used as the query to retrieve the example, and error types of the bold edits are assigned by ERRANT.}
\label{tbl:example}
\end{table*}

We analyzed the accuracy of EB-GEC for different error types to investigate the effect of error type on EB-GEC performance.
We used ERRANT to evaluate the accuracy of EB-GEC for each error type on the FCE-test. %\footnote {We cannot analyze error types of W\&I due to the unavailability of the gold sentences of its test data.} 

\autoref{tbl:error_type} shows three error types selected as having the most significant increase and decrease in accuracy for EB-GEC compared to the vanilla GEC. 
The three error types with the largest increases were preposition (\texttt{PREP}; e.g. \textit{I think we should book \underline{at/\hspace{0.3cm}} the Palace Hotel .}), punctuation error (\texttt{PUNCT}; e.g. \textit{Yours \underline{./\hspace{0.3cm}} sincerely ,}), and article error (\texttt{DET}; e.g. \textit{That should complete 
\underline{that/an} amazing day .}).
The three error types with the largest decreases are adjective conjugation error (\texttt{ADJ:FORM};  e.g. \textit{I was very \underline{please/pleased} to receive your letter .}), adjective error (\texttt{ADJ}; e.g. \textit{The adjoining restaurant is very \underline{enjoyable/good} as well .}), and spelling error (\texttt{SPELL}; e.g. \textit{Pusan Castle is \underline{locted/located} in the South of Pusan .}).

We concluded the following findings from these results.
Error types with the largest increase in accuracy have a limited number of tokens used for the edits compared to those with the largest decreases in accuracy (namely, error types referring to adjectives and nouns).
Furthermore, these error types are the most frequent errors in the datastore, (excluding the unclassified error type annotated as \texttt{OTHER}), and the datastore sufficiently covers such edits.
Contrary to the error types with improved accuracy, \texttt{ADJ} and \texttt{SPELL} have a considerable number of tokens used in edits, and they are not easy to cover sufficiently in a datastore.
Moreover, \texttt{ADJ:FORM} is the second least frequently occurring error type in the datastore, and we believe such examples cannot be covered sufficiently.
These results show that EB-GEC improves the accuracy of error types that are easily covered by examples, as there are fewer word types rarely used for edits and they are better presented in datastore.
Furthermore, the results show that the accuracy deteriorates for error types that are difficult to cover, such as word types used for edits and infrequent error types in the datastore.

We investigated the characteristics of the EB-GEC examples by comparing specific examples for each error type with those from token-based and BERT-based retrieval.
\autoref{tbl:example} shows examples of Token-based retrieval, BERT-based retrieval and EB-GEC for the top three error types (\texttt{PREP}, \texttt{PUNCT} and \texttt{DET}) with accuracy improvement in EB-GEC.
Token-based retrieval showed that the tokens in the edits are consistent, including ``\textbf{\textit{\underline{in/at}}}'', ``\textbf{\textit{\underline{\hspace{0.3cm}/,}}}'', and ``\textbf{\textit{\underline{\hspace{0.3cm}/a}}}''.
However, only surface information is used, and context is not considered.
So such unrelated examples are not useful for language learners.
BERT-based retrieval presented the same examples as EB-GEC for \texttt{PREP} and \texttt{PUNCT} error types, and the label for human evaluation was also 1.
However, the last example is influenced by the context rather than the correction and so presents an irrelevant example, labeled 0 by human evaluation.
This indicates that BERT-based retrieval overly focuses on context, resulting in examples related to the overall output but unrelated to the edits.
Conversely, EB-GEC is able to present examples in which the editing pair tokens are consistent for all corrections.
Furthermore, the contexts were similar to those of the input/output, for example ``\textit{purchase them \textbf{\underline{in/at}} reasonable \underline{price/prices}}'', ``\textit{\underline{for/For} example \textbf{\underline{\hspace{0.3cm}/,}}}'' and ``\textit{it takes \textbf{\underline{\hspace{0.3cm}/a}} long time to}'', and all the examples were labeled 1 during human evaluation.
This demonstrates that EB-GEC retrieves the most related examples that are helpful for users.

\section{Related Work}

\subsection{Example Retrieval for Language Learners}
There are example search systems that support language learners by finding examples. Before neural-based models, examples were retrieved and presented by surface matching~\cite{2008,yen-etal-2015-writeahead}.
\citet{arai-etal-2019-grammatical,arai2020example} proposed to combine Grammatical Error Detection (GED) and example retrieval to present both grammatically incorrect and correct examples of essays written by Japanese language learners.
This study showed that essay quality was improved by providing examples.
Their method is similar to EB-GEC in that it presents both correct and incorrect examples but incorporates example search systems for GED rather than the GEC.
Furthermore, the example search systems search for examples independently of the model.
Contrastingly, EB-GEC presents more related examples as shown in Section \ref{sec:result}.
%Still, it differs from the proposed method in that it focuses on detection rather than correction. 
%In their method, grammatical error detection, and example retrieval are independent and do not address the interpretability of the model.
%It is possible to apply such example search systems to the GEC model outputs to retrieve and present similar examples.
%On the other hand, the purpose of EB-GEC is to show that the interpretability of the GEC model is useful for learners, and using the examples retrieved by systems independent of the GEC model cannot be regarded as interpretability of the GEC model.

\citet{cheng-nagase-2012-example} developed a Japanese example-based system that retrieves examples using dependency structures and proofread texts.
Proofreading is a task similar to GEC because it also involves correcting grammatical errors.
However, this method also does not focus on using examples to improve interpretability.

\subsection{Explanation for Language Learners}

There is a feedback comment generation task~\cite{nagata-2019-toward} that can generate useful hints and explanations for grammatical errors and unnatural expressions in writing education.
\citet{nagata-etal-2020-creating} used a grammatical error detection model~\cite{kaneko-etal-2017-grammatical,kaneko2019multi} and neural retrieval-based method for prepositional errors.
The motivation of this study was similar to ours, that is, to help language learners understand grammatical errors and unnatural expressions in an interpretable way.
On the other hand, EB-GEC supports language learners using examples from the GEC model rather than using feedback.
%Furthermore, their study focuses only on prepositions.
%On the other hand, our study covers a wide range of errors.

\subsection{Example Retrieval in Text Generation}

Various previous studies have used neural network models to retrieve words, phrases, and sentences for use in prediction.
\citet{nagao1984framework} proposed an example-based MT to translate sequences by analogy.
This method has been extended to a variety of other methods for MT~\cite{sumita-iida-1991-experiments,10.1145/1113308.1113310,van2007memory,stroppa-etal-2007-exploiting,van2009extending,haque-etal-2009-using}.
In addition, the example-based method has been used for summarization~\cite{makino-yamamoto-2008-summarization} and paraphrasing~\cite{ohtake-yamamoto-2003-applicability}.
These studies were performed before neural networks were in general use, and the examples were not used to solve the neural network black box as was done in this study.

In neural network models, methods using examples have been proposed to improve accuracy and interpretability during inference.
\citet{gu2018search} proposed a model that during inference retrieves parallel sentences similar to input sentences and generates translations by the retrieved parallel sentences.
\citet{zhang-etal-2018-guiding} proposed a method that, during inference, retrieves parallel sentences where the source sentences are similar to the input sentences and weights the output containing $n$-grams of the retrieved sentence pairs based on the similarity between the input sentence and the retrieved source sentence.
These methods differ from EB-GEC using $k$NN-MT in that they retrieve examples via surface matching, as done in baseline token-based retrieval.
Moreover, these studies do not focus on the interpretability of the model.
%Since EB-GEC does not depend on the method of example retrieval, these methods can also be used.

Several methods have been proposed to retrieve examples using neural model representations and consider them for prediction.
\citet{Khandelwal2020Generalization,khandelwal2021nearest} proposed the retrieval of similar examples using the nearest neighbor examples of pre-trained hidden states during inference and to complement the output distributions of the language model and machine translation with the distributions of these examples.
\citet{NEURIPS2020_6b493230} combined a pre-trained retriever with a pre-trained encoder-decoder model and fine-tuned it end-to-end.
For the input query, they found the top-$k$ documents and used them as a latent variable for final prediction.
\citet{guu2020realm} first conducted an unsupervised joint pre-training of the knowledge retriever and knowledge-augmented encoder for the language modeling task, then fine-tuned it using a task of primary interest, with supervised examples.
The main purpose of these methods was to improve the accuracy using examples, and whether the examples were helpful for the users was not verified.
Conversely, our study showed that examples for the interpretability in GEC could be helpful for real users.

\section{Conclusion}

We introduced EB-GEC to improve the interpretability of corrections by presenting examples to language learners.
The human evaluation showed that the examples presented by EB-GEC supported language learners' decision to accept corrections and improved their understanding of the correction results.
Although existing interpretive methods using examples have not verified if examples are helpful for humans, this study demonstrated that examples were helpful for learners using GEC.
In addition, the results of the GEC benchmark showed that EB-GEC could predict corrections more accurately or comparably to its vanilla counterpart.
%Experimental results show that EB-GEC improves accuracy by using examples while also improving interpretability.
%By proposing EB-GEC, we hope that GEC will focus not only on accuracy but also on interpretability in the future.

Future work would include investigations of whether example presentation is beneficial for learners with low language proficiency.
In addition, we plan to improve the datastore coverage by using pseudo-data~\cite{xie-etal-2018-noising} and weight low frequency error types to present diverse examples.
We explore whether methods to improve accuracy and diversity~\cite{chollampatt-ng-2018-multilayer,kaneko-etal-2019-tmu,hotate-etal-2019-controlling,hotate-etal-2020-generating} are effective for EB-GEC.

\section*{Acknowledgements}

This paper is based on results obtained from a project, JPNP18002, commissioned by the New Energy and Industrial Technology Development Organization (NEDO).
We thank Yukiko Konishi, Yuko Unzai, and Naoko Furuya for their help with our experiments.

\bibliography{custom}
\bibliographystyle{acl_natbib}

\end{document}